\documentclass{article} % For LaTeX2e

\PassOptionsToPackage{numbers, compress}{natbib}
\usepackage[final]{neurips_2019}

\usepackage{hyperref}
\usepackage{url}
\usepackage{siunitx}
\usepackage{graphicx}
\usepackage{floatrow}
\usepackage{caption}
\usepackage{subcaption}
\usepackage{xcolor}

\newfloatcommand{capbtabbox}{table}[][\FBwidth]

\title{Knee Cartilage Segmentation Using Diffusion-Weighted MRI}

\author{Alejandra Duarte\thanks{Equal Contribution. Names in alphabetical order.} \thanks{Department of Radiology, New York University School of Medicine, New York, NY 10016, USA}  \\
\texttt{alejandra.duarte@nyulmc.org} \\
\And
Chaitra V. Hegde$^{*}$ \thanks{Center for Data Science, New York University, New York, NY 10011, USA} \\
\texttt{cvh255@nyu.edu} \\
\And
Aakash Kaku$^{* \ddagger}$ \\
\texttt{ark576@nyu.edu} \\
\And
Sreyas Mohan$^{*\ddagger}$ \\
\texttt{sm7582@nyu.edu} \\
\And
Jos\'e G. Raya$^{\dagger}$ \\
\texttt{jose.raya@nyulmc.org} \\
}

\begin{document}

\maketitle

\begin{abstract}
The integrity of articular cartilage is a crucial aspect in the early diagnosis of osteoarthritis (OA). Many novel MRI techniques have the potential to assess compositional changes of the cartilage extracellular matrix. Among these techniques, diffusion tensor imaging (DTI) of cartilage provides a simultaneous assessment of the two principal components of the solid matrix: collagen structure and proteoglycan concentration. DTI, as for any other compositional MRI technique, require a human expert to perform segmentation manually. The manual segmentation is error-prone and time-consuming ($\sim$ few hours per subject). We use an ensemble of modified U-Nets to automate this segmentation task. We benchmark our model against a human expert test-retest segmentation and conclude that our model is superior for Patellar and Tibial cartilage using dice score as the comparison metric. In the end, we do a perturbation analysis to understand the sensitivity of our model to the different components of our input. We also provide confidence maps for the predictions so that radiologists can tweak the model predictions as required. The model has been deployed in practice. In conclusion, cartilage segmentation on DW-MRI images with modified U-Nets achieves accuracy that outperforms the human segmenter. Code is available at \href{https://github.com/aakashrkaku/knee-cartilage-segmentation}{\tt{https://github.com/aakashrkaku/knee-cartilage-segmentation}}.
\end{abstract}

\section{Background}

Osteoarthritis (OA) is the most prevalent knee joint disease in the United States. OA eventually leads to chronic disability \citep{cdc}, making the early detection of OA very important. OA is characterized in its early stages by the degeneration of articular cartilage. To assess the integrity of the cartilage, its biochemical composition needs to be measured \citep{jose1,jose2}. Several compositional MRI techniques have been introduced that are sensitive to either proteoglycan (delayed Gadolinium-enhanced MRI [dGEMRIC], Sodium-MRI, glycosominoglycans chemical exchange saturation transfer [gagCEST]) or to a combination of components (T2 relaxation time, relaxation time in the rotating frame [T1$\rho$], magnetization transfer). Recently, DTI was introduced as a novel biomarker that can capture proteoglycan content and collagen structure simultaneously \citep{jose1,jose2,jose3}. A significant limitation of the use in the clinical routine of these advanced MRI techniques is the lengthy image processing time \citep{jose3}. To measure MRI parameters in cartilage, the cartilage needs to be segmented, which is usually done by a human expert, a trained Radiologist, and takes few hours to complete segmenting all cartilage plates (tibia, femur, patella) from a patient. This way of segmentation is not scalable, error-prone, and extremely slow. Deep learning-based models are successful in performing quick and accurate segmentation of brain tissues \cite{darts, slant}, tumor \cite{kamnitsas2015multi}, pancreas \cite{cai2016pancreas}, cardiac substructures \cite{avendi2016combined} and other biological parts. Here, we propose to use deep learning methods to automate the segmentation of knee cartilages. 

\section{Methodology}
\label{sec:methodology}
\begin{figure}
\begin{floatrow}
\ffigbox{%
  \includegraphics[width=2.2in, height = 1.1in]{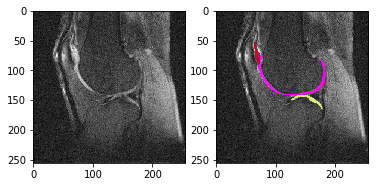}%
}{%
\caption{Example of a dataset with the ground truth articular cartilages segmentation marked in colors. Red: Patella, Yellow: Tibia, Pink: Femur}
\label{fig:sample image}
}
\capbtabbox{%
\begin{tabular}{lll}
\multicolumn{1}{c}{\bf Segment}  &\multicolumn{1}{c}{\bf Ave. Voxel Count} &\multicolumn{1}{c}{\bf \%}
\\ \hline \\
Femur         &1083 &1.659\%\\
Patella             &260&0.397\% \\
Tibia             &186&0.284\% \\
None             & 64007&97.66\%\\
Total             & 256 $\times$ 256 &100\%\\
\end{tabular}
% \label{sample-table}
}{%
  \caption{Distribution of labels for each segment}%
  \label{sample-table}
}
\end{floatrow}

\end{figure}

%% Dataset is part of the method not  the introduction. Thus I moved here the headign JR
%% Dataset Starts Here
We work with a dataset of diffusion-weighted MRI of knees of OA patients, which has 71 MRI scans for $36$ patients whose OA severity ranges from none (Kellgren-Lawrence [KL] score of 0) to severe (KL=3) \citep{klscore}. DTI images were acquired using a radial imaging spin-echo diffusion sequence (RAISED, TR/TE=1500/35 ms, matrix=256$\times$256, 15 slices, resolution=0.6$\times$0.6$\times$3~mm$^3$, b value=300 s/mm$^2$ , 6 directions, 105 spokes/image, acquisition time 17:50 min).  For a good segmentation, one should be able to distinguish between articular cartilage and fluid, which have similar voxel intensity in the diffusion-weighted images. We compute the mean diffusivity (MD) maps, and fractional anisotropy (FA) maps from these seven contrast images that help to resolve the issue of distinguishing cartilage from fluid \citep{jose1}. MD and FA from the seven contrast images make the input of an individual patient to be a matrix of size 256$\times$256$\times$15$\times$9. The ground truth of all cartilage plates (lateral and medial tibia, femur, and patella) labels associated with each diffusion-weighted map, were generated by a musculoskeletal radiologist. A subset of five images was resegmented by the radiologist to provide a benchmark for expert human coincidence (A sample map and segmentation are shown in Fig \ref{fig:sample image}).
%% I think is very important to add the re-segmentation since this was the argument to show that software was having similar accuracy to human JR.

We split the data set into a train (80\%), validation (10\%), and test (10\%) set, making sure that the subjects in these sets are disjoint. We preprocess each channel independently using min-max normalization to keep the range from $0-1$. In addition to this, to complement the limited data we have, the training set was augmented using random rotation ( \ang{-5} to \ang{5}) and random horizontal and vertical shift ($-10$ to $10$ pixels) of the images, which are perturbations that could happen during the procurement of MRI Images because of small movements of the patients. We did not use vertical or horizontal flipping as they are not natural perturbations in our setting.

\begin{figure}
\begin{floatrow}
\ffigbox{%
  \includegraphics[width=2.3in, height = 1.76in]{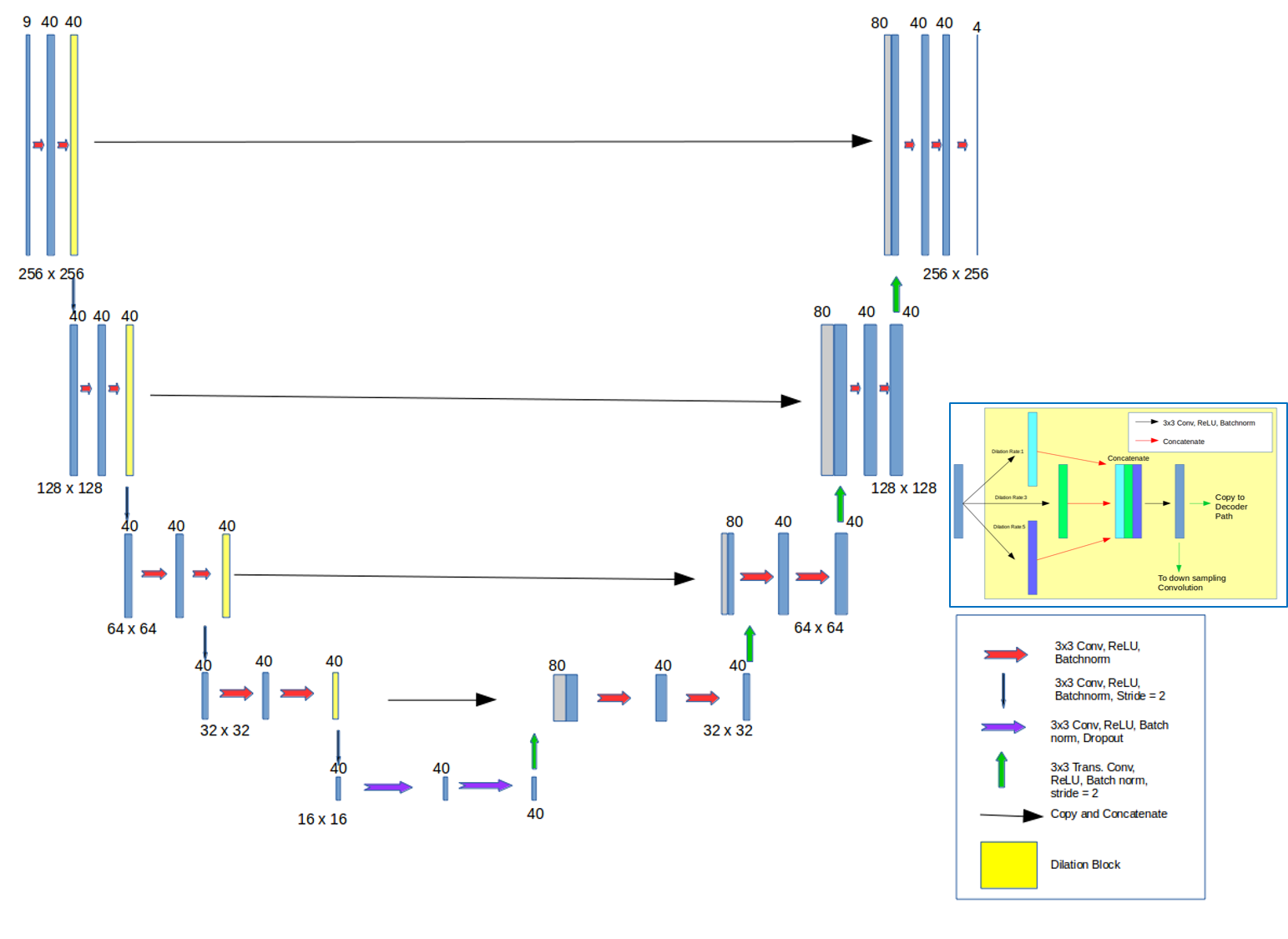}%
}{%
\caption{Architecture of dilated UNet. Dilation helps achieve a larger field of view.}
\label{fig:unet_dilated}
}
\capbtabbox{%
\begin{tabular}{llll}
\multicolumn{1}{c}{\bf Model}  &\multicolumn{1}{c}{\bf Femur} &\multicolumn{1}{c}{\bf Patella} &\multicolumn{1}{c}{\bf Tibia}
\\ \hline \\
3D U-Net \cite{cciccek20163d}& 0.618&0.697&0.378 \\
VNet \cite{vnet} & 0.625 & 0.567 & 0.467 \\
2D U-Net& 0.678&0.773&0.593\\
2D Dilated U-Net&0.670&0.771&0.621\\
\bf Ensemble of 2D U-Net \\ \bf and Dilated U-Net &0.689 &\bf 0.783&\bf 0.640\\
\\ \hline \\
Human Expert \\(Re-segmentation) & \bf 0.711 &0.743 & 0.629\\
\\ \hline \\
\end{tabular}
% \label{tab:performance}
}{
  \caption{Performance of different models in terms of Dice Score on the validation set. We see that our ensemble model matches the performance of a human expert. Performance of the ensemble model on the test set is as follows: \textbf{F: $0.691$, P: $0.778$, T: $0.681$} }
  \label{tab:performance}
}
\end{floatrow}

\end{figure}

We model the segmentation task as a $4$-class classification problem - $3$ tissues of interest and no tissue. Since our dataset is highly imbalanced (Table \ref{sample-table}), we optimize the Weighted Dice Loss \citep{wcel}, where we followed \citep{wcel} to estimate the weights. We use Adam optimizer \citep{adam} with a learning rate of $1e-4$ to optimize the model parameters. We assume that each spatial slice is independent of others, or in other words, the location of the slice in the 3-D cube does not matter. We validated this assumption by training a 3D convolution, which did not change our results (Table \ref{tab:performance}). 

We train the original UNet proposed in \citep{unet} with $40$ channels and a version of UNet with dilated convolutions (Fig \ref{fig:unet_dilated}). We analyze the predictions made by each of the models we trained and find that the nature of the error these models make is very different from each other (Fig \ref{fig:error_analysis}).  Therefore, as our final model, we ensemble these two models using a three-layered CNN, which has $8$ input channels ($4$ from the output of each network) and achieved better performance (Table \ref{tab:performance}).

% \begin{figure}%
%     \centering
%     \subfloat[label 1]{\includegraphics[width=5cm]{pred_4.png}%
% \caption{Case where the model was more conservative than the human expert. Left: Original Image, Center: Ground Truth, Right: Model Prediction. Femur = Pink, Patella = Red, Tibia = Yellow}
% \label{fig:pred_1}}%
%     \qquad
%     \subfloat[label 2]{{\includegraphics[width=5cm]{cert_image.png}%
% \caption{A confidence map for two example validation images can be seen. The circled stray voxels are incorrectly classified as one of the cartilage. It can be seen that incorrect voxels have low confidence}
% \label{fig:cert_image}}}%
%     \caption{2 Figures side by side}%
%     \label{fig:example}%
% \end{figure}
\begin{figure}[ht]
\centering
\includegraphics[width= 0.9 \textwidth]{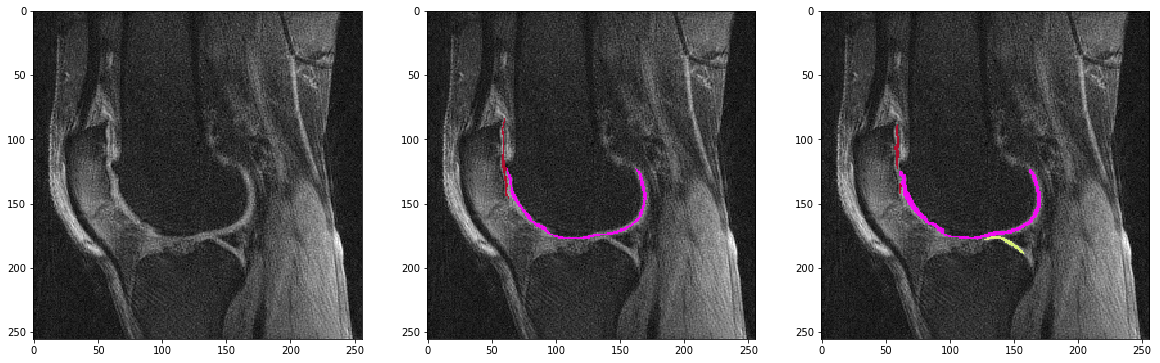}%
\caption{Example of an output where model correctly predicted a cartilage that was missed in the ground truth segmentation by the radiologist. Left: Original Image, Center: Ground Truth, Right: Model Prediction. Femur = Pink, Patella = Red, Tibia = Yellow. See Fig~\ref{fig:other_knee_examples} for more examples.}
\label{fig:pred_1}
\end{figure}

\begin{figure}
\centering
\includegraphics[width=2.8in, height = 1.4in]{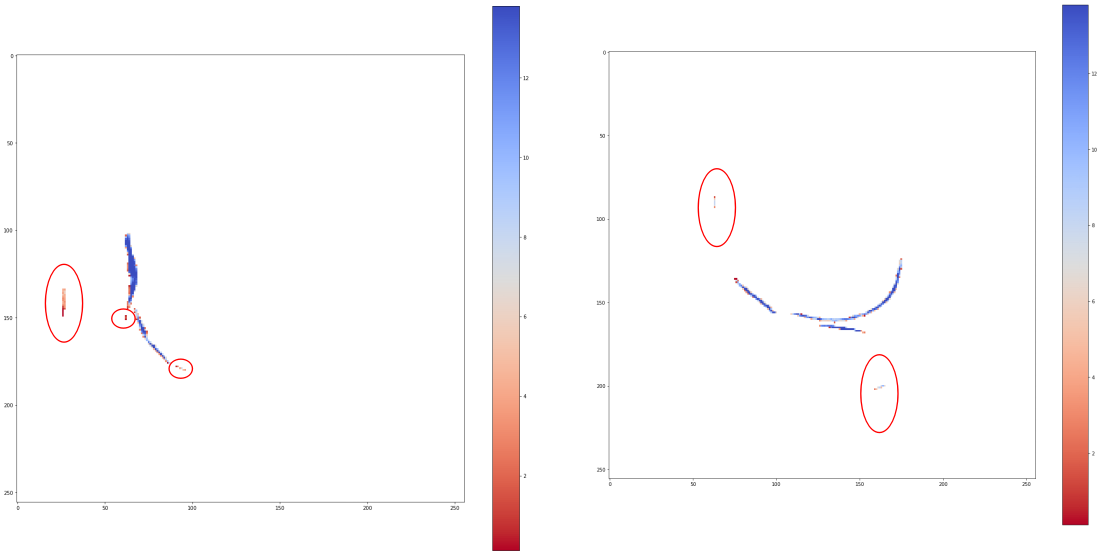}%
\caption{A confidence map for two sample images can be seen. The circled stray pixels are incorrectly classified as one of the cartilage. It can be seen that incorrect pixels have low confidence.}
\label{fig:cert_image}
\end{figure}
% \begin{figure}
% \centering
% \begin{minipage}{.5\textwidth}
%   \centering
%   \includegraphics[width=7cm]{pred_2.png}%
%   \caption{Case where the model was more conservative than the human expert. Left: Original Image, Center: Ground Truth, Right: Model Prediction. Femur = Pink, Patella = Red, Tibia = Yellow}
%   \label{fig:pred_1}
% \end{minipage}%
% \begin{minipage}{.5\textwidth}
%   \centering
%   \includegraphics[width=7cm]{cert_image.png}%
%   \caption{A confidence map for two example validation images can be seen. The circled stray voxels are incorrectly classified as one of the cartilage. It can be seen that incorrect voxels have low confidence}
%   \label{fig:cert_image}
% \end{minipage}
% \end{figure}

\section{Analysis}
\label{sec:analysis}
\textbf{Estimating Human Performance}:  Due to the low contrast of DW-MRI images, the estimation of the correct label for each pixel is inherently noisy. This is corroborated by the observation that when the same human expert segments some of the images again, the dice score calculated on these re segmented images is much less than $1$ (Table \ref{tab:performance}). Further, we visually observe that the distribution of dice score (considering the first segmentation as ground truth) for both the human expert and our model for all the images that were re-segmented is very similar, implying that, the nature of error made by the human expert and our model is similar (Fig \ref{fig:dice_distribution}).

\textbf{Confidence Map for Predictions}: Fig \ref{fig:pred_1} shows an instance where our model predicted the existence of a tissue that was absent in the ground truth. The radiologist confirmed that this was mislabelled. Since ground truth segmentations were noisy, we calculate a confidence map to guide on the certainty of the model prediction. We calculate the log of odds ratio, $log(\frac{p_{max}}{1-p_{max}})$, where $p_{max}$ is the maximum probability of all $4$ classes, for each voxel to quantify our confidence in the prediction of this voxel. We present this confidence map along with our segmentation to the radiologists (Fig \ref{fig:cert_image}). These confidence maps are particularly useful, as we are working with \emph{diffusion weighted MRI}, which has poor contrast between cartilage and surrounding tissues making the labels around cartilage boundaries ambiguous and prone to error. 

\textbf{Sensitivity of Inputs for Trained Model}: We selectively set one of the channels to zero and studied model performance (Dice score) to understand how much the model relies on each of the input channels. The performance drops drastically with the removal of any channel, but more when MD and FA maps are zeroed out. Further, we found that the performance was almost unchanged when we permute the seven contrast maps, implying that the model gave similar weights to all the seven contrast maps and treated the calculated MD and FA maps independently. Similar weights or averaging over contrasts help the model to be invariant to the addition of noise to inputs.  

\section{Conclusion}
We built a deep learning system to perform semi-automatic segmentation of knee cartilages from diffusion-weighted MRI in less than a minute on a non-GPU machine. We showed that the Unet-based models perform similar to a human expert. Further, this model is deployed for clinical trials, and radiologists are currently using it to characterize the progression of OA disease. We find that the confidence maps are particularly helpful in determining which pixels to concentrate on for the radiologists. In practice, we see that the segmentation can be used out of the box in most of the cases without any manual corrections.

\section*{Acknowledgement}
We thank Dr. Bonnie K Ray, Dr. David Rosenberg, Dr. Narges Razavian and Dr. Cem Deniz for useful discussions, and guidance. Research reported in this manuscript was supported by the National Institute of Arthritis and Musculoskeletal and Skin Diseases (NIAMS) of the National Institute of Health (NIH) under award number and RO1AR067789. The content is solely the responsibility of the authors and does not necessarily represent the official views of the NIH.

\bibliographystyle{acm}
\bibliography{iclr2017_conference.bib}

\begin{thebibliography}{10}

\bibitem{avendi2016combined}
{\sc Avendi, M., Kheradvar, A., and Jafarkhani, H.}
\newblock A combined deep-learning and deformable-model approach to fully
  automatic segmentation of the left ventricle in cardiac mri.
\newblock {\em Medical image analysis 30\/} (2016), 108--119.

\bibitem{cai2016pancreas}
{\sc Cai, J., Lu, L., Zhang, Z., Xing, F., Yang, L., and Yin, Q.}
\newblock Pancreas segmentation in mri using graph-based decision fusion on
  convolutional neural networks.
\newblock In {\em International Conference on Medical Image Computing and
  Computer-Assisted Intervention\/} (2016), Springer, pp.~442--450.

\bibitem{cdc}
{\sc CDC}.
\newblock Prevalence of disabilities and associated health conditions among
  adults--united states 1999.
\newblock {\em Centers for Disease Control and Prevention (CDC) 50\/} (2001),
  149.

\bibitem{cciccek20163d}
{\sc {\c{C}}i{\c{c}}ek, {\"O}., Abdulkadir, A., Lienkamp, S.~S., Brox, T., and
  Ronneberger, O.}
\newblock 3d u-net: learning dense volumetric segmentation from sparse
  annotation.
\newblock In {\em International conference on medical image computing and
  computer-assisted intervention\/} (2016), Springer, pp.~424--432.

\bibitem{slant}
{\sc Huo, Y., Xu, Z., Aboud, K., Parvathaneni, P., Bao, S., Bermudez, C.,
  Resnick, S.~M., Cutting, L.~E., and Landman, B.~A.}
\newblock Spatially localized atlas network tiles enables 3d whole brain
  segmentation from limited data.
\newblock In {\em International Conference on Medical Image Computing and
  Computer-Assisted Intervention\/} (2018), Springer, pp.~698--705.

\bibitem{darts}
{\sc Kaku, A., Hegde, C.~V., Huang, J., Chung, S., Wang, X., Young, M., Lui,
  Y.~W., and Razavian, N.}
\newblock Darts: Denseunet-based automatic rapid tool for brain segmentation.
\newblock {\em arXiv preprint arXiv:1911.05567\/} (2019).

\bibitem{kamnitsas2015multi}
{\sc Kamnitsas, K., Chen, L., Ledig, C., Rueckert, D., and Glocker, B.}
\newblock Multi-scale 3d convolutional neural networks for lesion segmentation
  in brain mri.
\newblock {\em Ischemic stroke lesion segmentation 13\/} (2015), 46.

\bibitem{klscore}
{\sc Kellgren, J., and Lawrence, J.}
\newblock Radiological assessment of osteo-arthrosis.
\newblock {\em Annals of the rheumatic diseases 16}, 4 (1957), 494.

\bibitem{adam}
{\sc Kingma, D.~P., and Ba, J.}
\newblock Adam: {A} method for stochastic optimization.
\newblock {\em CoRR abs/1412.6980\/} (2014).

\bibitem{vnet}
{\sc Milletari, F., Navab, N., and Ahmadi, S.}
\newblock V-net: Fully convolutional neural networks for volumetric medical
  image segmentation.
\newblock {\em CoRR abs/1606.04797\/} (2016).

\bibitem{jose1}
{\sc Raya, J., Horng, A., Dietrich, O., Krasnokutsky, S., Beltran, L.~S.,
  Storey, P., Reiser, M.~F., Recht, M.~P., Sodickson, D.~K., Glaser, C., and
  et~al.}
\newblock Articular cartilage: In vivo diffusion-tensor imaging.
\newblock {\em Radiology 262}, 2 (2012), 550–559.

\bibitem{jose3}
{\sc Raya, J., Melkus, G., Adam-Neumair, S., Dietrich, O., Mützel, E., Kahr,
  B., Reiser, M.~F., Jakob, P.~M., Putz, R., Glaser, C., and et~al.}
\newblock Change of diffusion tensor imaging parameters in articular cartilage
  with progressive proteoglycan extraction.
\newblock {\em Investigative Radiology 46}, 6 (2011), 401–409.

\bibitem{jose2}
{\sc Raya, J., Melkus, G., Adam-Neumair, S., Dietrich, O., Mützel, E., Reiser,
  M.~F., Putz, R., Kirsch, T., Jakob, P.~M., Glaser, C., and et~al.}
\newblock Diffusion-tensor imaging of human articular cartilage specimens with
  early signs of cartilage damage.
\newblock {\em Radiology 266}, 3 (2013), 831–841.

\bibitem{unet}
{\sc Ronneberger, O., Fischer, P., and Brox, T.}
\newblock U-net: Convolutional networks for biomedical image segmentation.
\newblock {\em CoRR abs/1505.04597\/} (2015).

\bibitem{wcel}
{\sc Sudre, C.~H., Li, W., Vercauteren, T., Ourselin, S., and Cardoso, M.~J.}
\newblock Generalised dice overlap as a deep learning loss function for highly
  unbalanced segmentations.
\newblock {\em CoRR abs/1707.03237\/} (2017).

\end{thebibliography}

\appendix

\section{Additional Results}
\begin{figure}
\centering
\includegraphics[width = \textwidth]{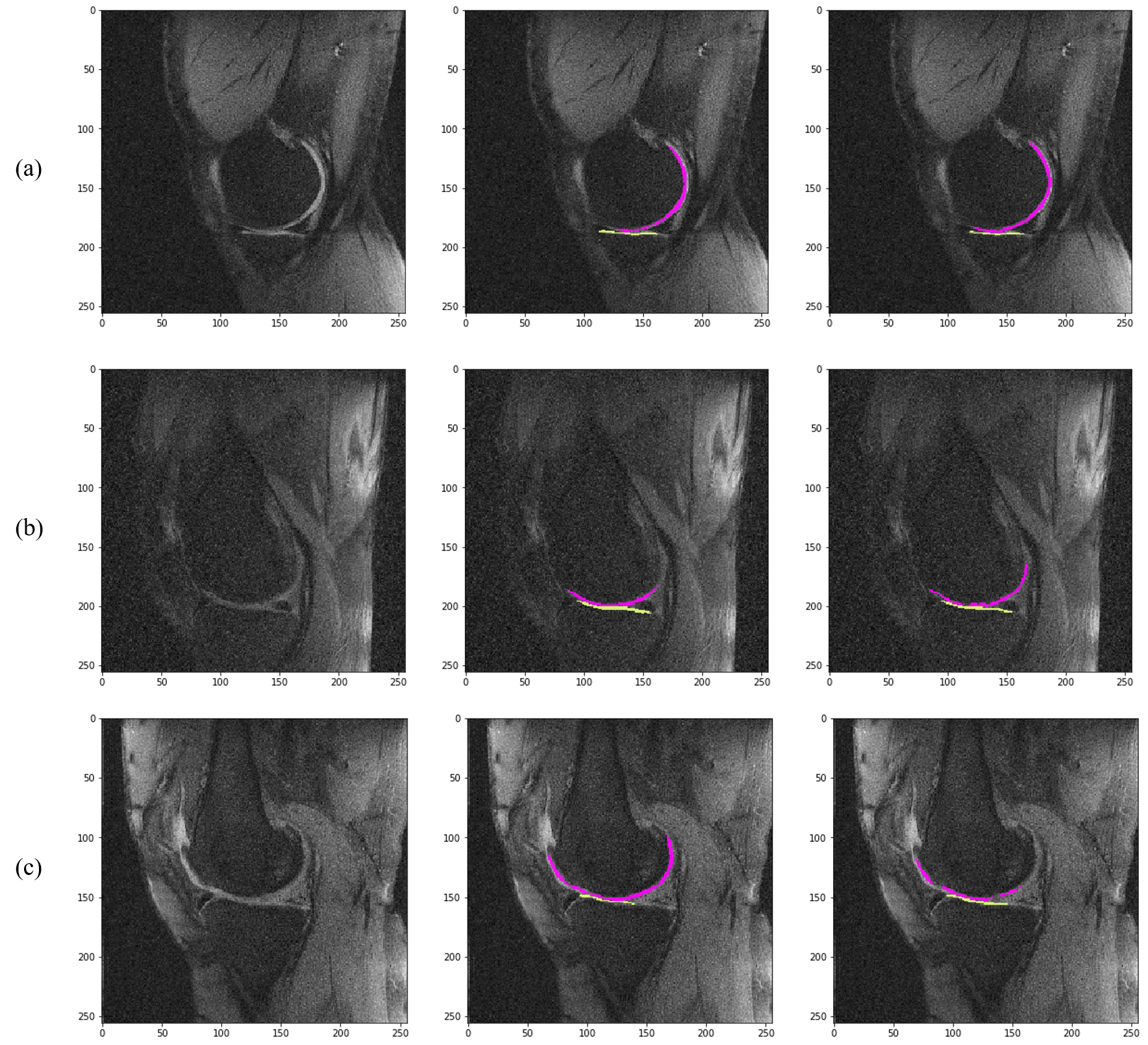}%
\caption{ Example of an output (a) where the model correctly predicted the ground truth cartilage (b) where the model was more conservative than the radiologist (c) where the radiologist was more conservative than the model. Left: Original Image, Center: Ground Truth, Right: Ensemble Model Prediction. Femur = Pink, Patella = Red, Tibia = Yellow}
\label{fig:other_knee_examples}
\end{figure}

\begin{figure}
\centering
\includegraphics[width = \textwidth]{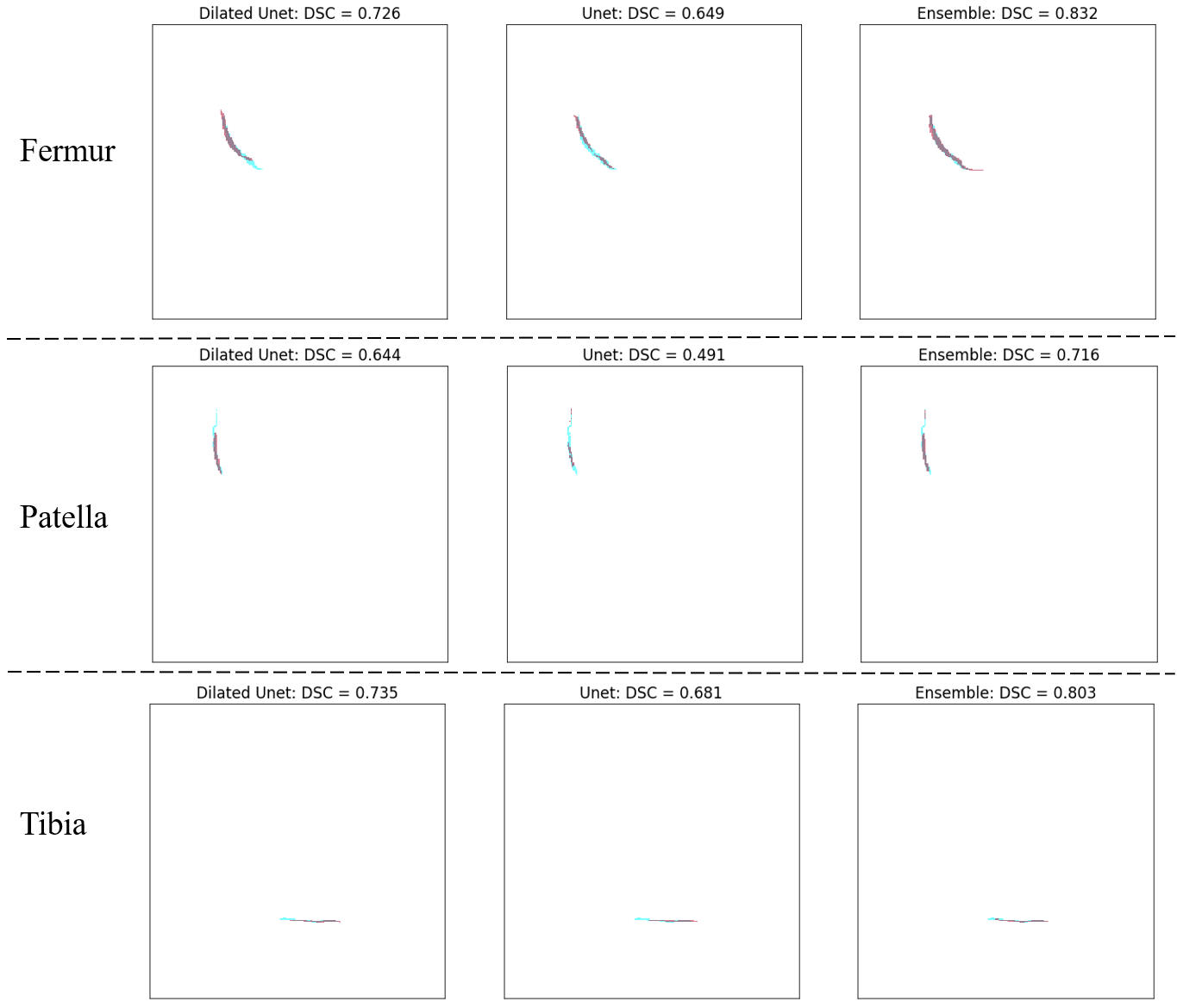}%
\caption{Each image shows the ground truth segmentation (blue) overlayed on top of the prediction from the model (red). We see that the Dilated UNet model (left) and the UNet (center) makes different kind of errors and ensembling them (right) can correct such errors.}
\label{fig:error_analysis}
\end{figure}

\begin{figure}
\centering
\includegraphics[width = \textwidth]{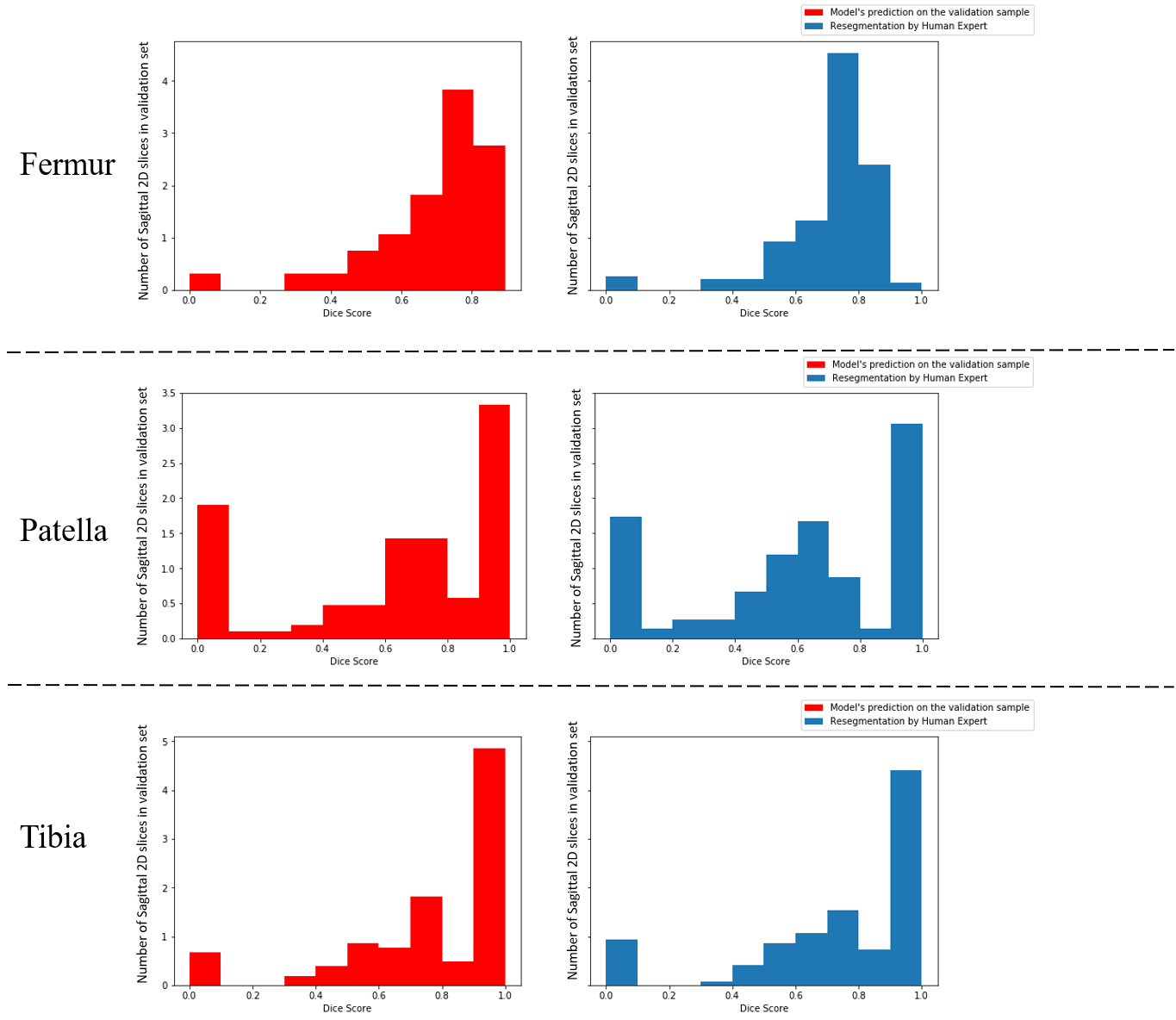}%
\caption{Distribution of dice score from model predictions (left) and from human experts (right) on re-segmented images (Section \ref{sec:analysis}). The initial segmentation produced by the human expert is used as the ground truth for calculating the dice score. Visually, one can observe that the distribution of the dice score for both the model and human expert is very similar to the re-segmented images, and hence, one could infer that the nature of error made by both the human expert and model is similar. }
\label{fig:dice_distribution}
\end{figure}

In this section, we provide additional results of our computational experiments:
\begin{itemize}
    \item Fig \ref{fig:other_knee_examples} shows specific instances of the model being conservative or aggressive relative to the ground truth predictions. 
    \item Fig \ref{fig:error_analysis} shows particular examples showing that the U-Net and Dilated-UNet models discussed in section \ref{sec:methodology} produces a different kind of errors. An ensemble model of both of these networks can correct for it and produce better segmentation results.
    \item Fig \ref{fig:dice_distribution} shows the distribution of dice scores on re-segmented images fo both the human expert and our model. The segmentation produced by the human expert during the first time is chosen as the ground truth for dice score calculations. Visually, the distributions look similar, and hence, one can infer that the nature of error made by the human expert and the model is similar. 
\end{itemize}

\end{document}